%% file: main.tex
\documentclass{article}

\PassOptionsToPackage{numbers, compress}{natbib}

\usepackage[preprint]{neurips_2025}

\usepackage{algorithm}
\usepackage{algpseudocode}
\usepackage[utf8]{inputenc} 
\usepackage[T1]{fontenc}    
\usepackage[pagebackref,breaklinks,colorlinks]{hyperref}
\usepackage[toc,page]{appendix}
\usepackage{url}            
\usepackage{booktabs}       
\usepackage{amsfonts}       
\usepackage{nicefrac}       
\usepackage{microtype}      
\usepackage{xcolor}         
\usepackage{graphicx}

\usepackage{amsmath}
\usepackage{enumitem}
\usepackage{wrapfig}

\title{Personalize Your Gaussian: Consistent 3D Scene Personalization from a Single Image}

\def\mystrut{\rule{0pt}{1.0\normalbaselineskip}}
\author{
\begin{tabular}{@{}c}
Yuxuan Wang$^{1}$\quad  
Xuanyu Yi$^{1}$\quad  
Qingshan Xu$^{1}$\quad  
Yuan Zhou$^{1}$\quad  
Long Chen$^{2}$\quad  
Hanwang Zhang$^{1}$\quad  
\\
\end{tabular}\\
$^1$Nanyang Technological University \mystrut \\
$^2$Hong Kong University of Science and Technology
}

\input{define}
\begin{document}

\maketitle

\begin{figure}[h]
\centering
\includegraphics[width=0.99\linewidth]{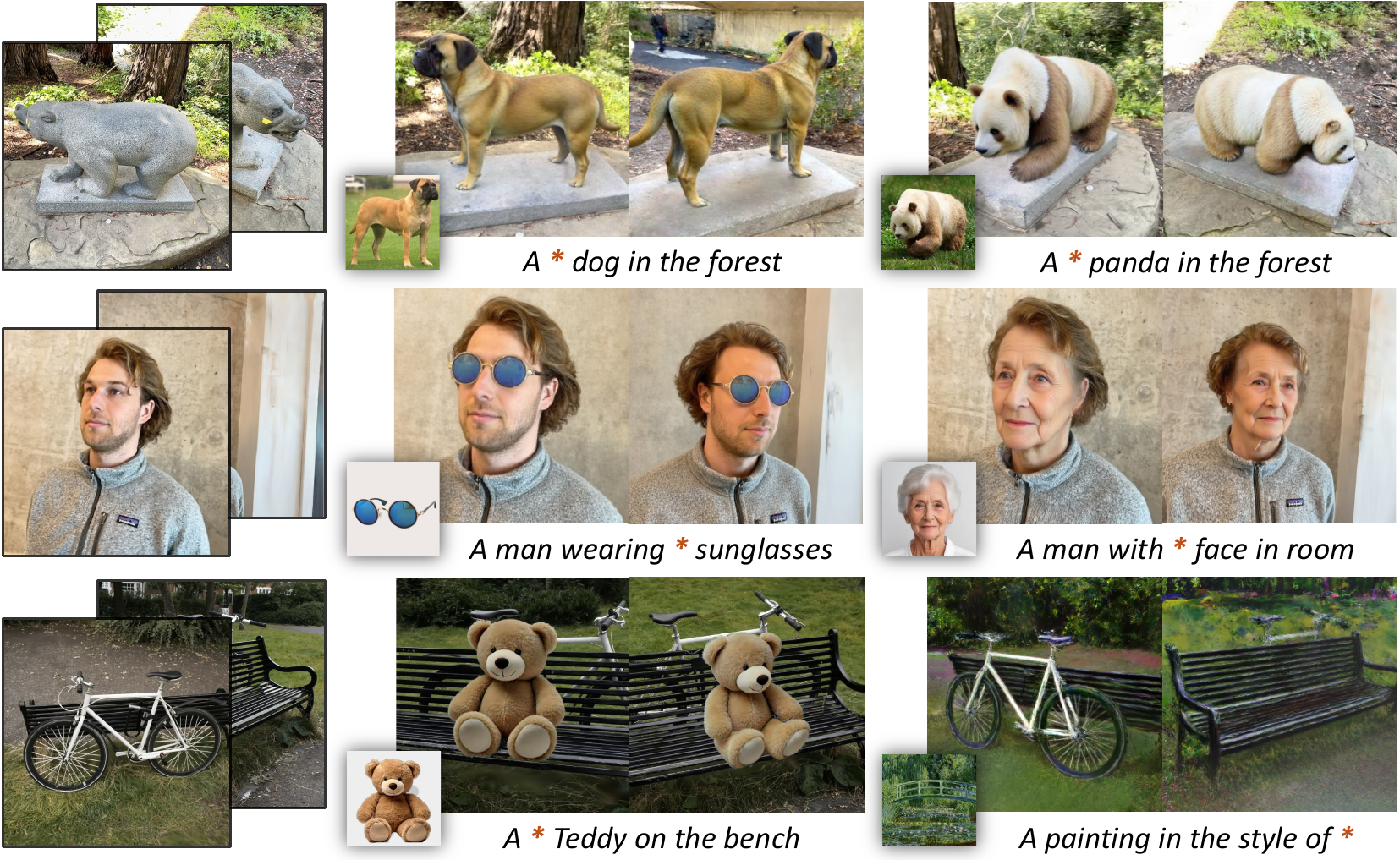}
\caption{Given a source 3DGS scene and a single reference image, $\method$ enables high-quality personalization by editing a user-specified region (e.g., the \textit{bear}, \textit{man’s eye}, \textit{man’s face}, \textit{bench-top}, \textit{entire scene}) to match the reference appearance, supporting replacement, adding, and style transfer.}
\label{fig:1}
\end{figure}

\begin{abstract}

Personalizing 3D scenes from a single reference image enables intuitive user-guided editing, which requires achieving both multi-view consistency across perspectives and referential consistency with the input image.
However, these goals are particularly challenging due to the viewpoint bias caused by the limited perspective provided in a single image.
Lacking the mechanisms to effectively expand reference information beyond the original view, existing methods of image-conditioned 3DGS personalization often suffer from this viewpoint bias and struggle to produce consistent results.
Therefore, in this paper, we present \textbf{Consistent Personalization for 3D Gaussian Splatting ($\method$)}, a framework that progressively propagates the single-view reference appearance to novel perspectives.
In particular, $\method$ integrates pre-trained image-to-3D generation and iterative LoRA fine-tuning to extract and extend the reference appearance, and finally produces faithful multi-view guidance images and the personalized 3DGS outputs through a view-consistent generation process guided by geometric cues.
Extensive experiments on real-world scenes show that our $\method$ effectively mitigates the viewpoint bias, achieving high-quality personalization that significantly outperforms existing methods.

\end{abstract}

\input{sec/1_intro}
\input{sec/2_related_work}
\input{sec/3_method}
\input{sec/4_experiment}
\input{sec/5_conclusion}

\clearpage

{
    \small
    \bibliographystyle{unsrtnat}
    \bibliography{main}
}

\clearpage

\appendix
\begin{appendices}

\noindent
The \textit{Appendix} is organized as follows:
\begin{itemize}[leftmargin=*]
    \item \textbf{Appendix~\ref{supp_sec:imple}:} provides \textbf{more details of implementation}, including a step-by-step pipeline demonstration, the coarse asset integration, the extension of LoRA training set, and more details of our user study.
    \item \textbf{Appendix~\ref{supp_sec:exp}:} further provides \textbf{additional experimental results}, comparisons, and analyses, including extended visualizations and more in-depth quantitative ablation studies.
    \item \textbf{Appendix~\ref{supp_sec:discuss}:} provides \textbf{more discussions} on the limitation and potential societal impacts with solutions of our $\method$.
\end{itemize}

\input{supp_sec/1_more_imp}
\input{supp_sec/2_ext_exp}
\input{supp_sec/3_other_discuss}

\begin{figure}[ht]
\centering
\includegraphics[width=1.\linewidth]{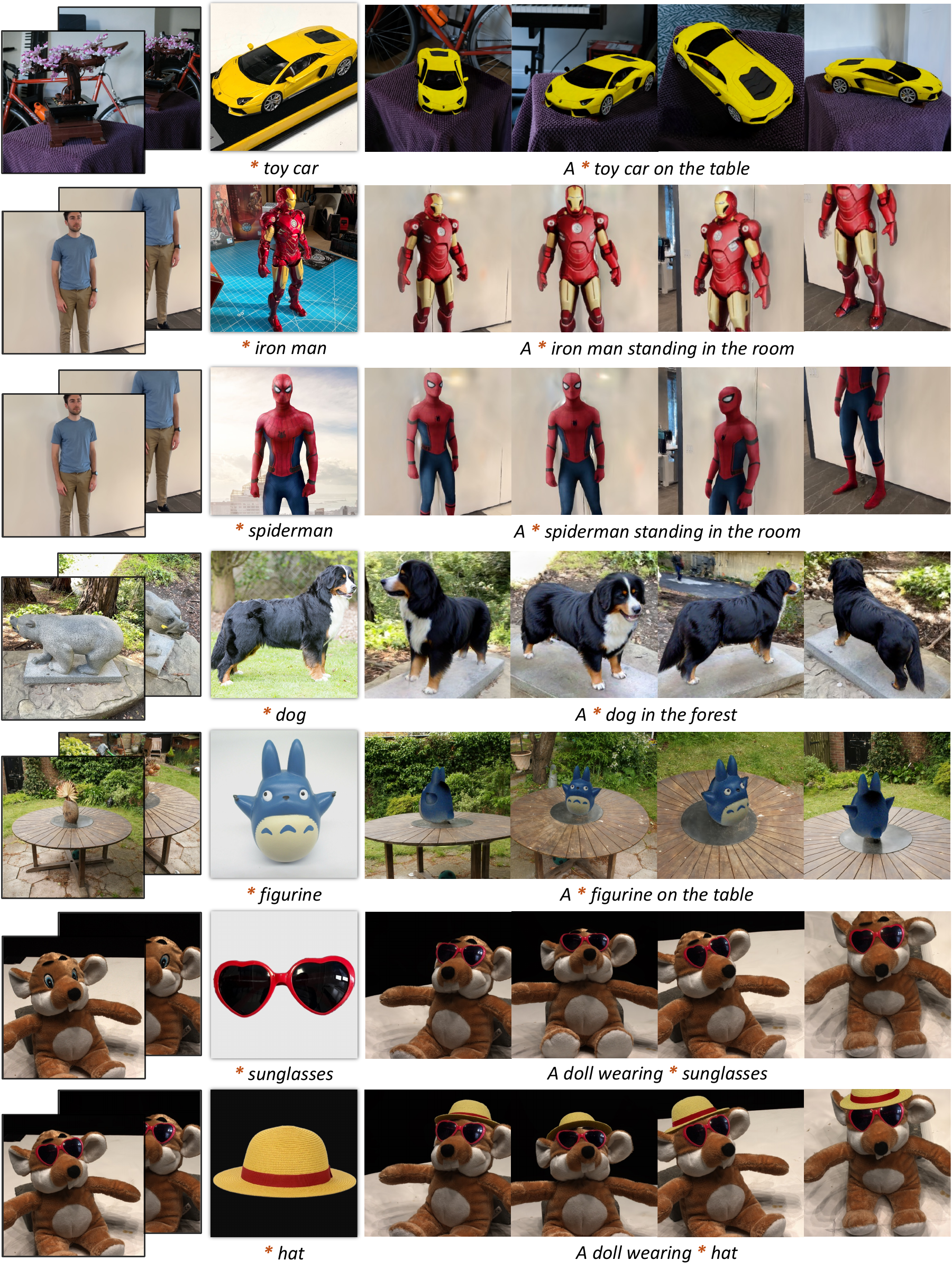}
\caption{Supplementary personalization result of our CP-GS, demonstrating high-quality 3DGS scene
customization that faithfully align with the reference image across various scenarios.}
\label{fig:9}
\end{figure}

\end{appendices}

\end{document}

%% file: define.tex
\newcommand{\method}{\text{CP-GS}}
\newcommand{\gsrc}{\mathcal{G}^\text{src}}
\newcommand{\gedt}{\mathcal{G}^\text{edit}}
\newcommand{\gcoar}{\mathcal{G}^\text{coar}}
\newcommand{\G}{\mathcal{G}}

\newcommand{\upix}{\mathbf{u}}
\newcommand{\vpix}{\mathbf{v}}
\newcommand{\Iedt}{\mathcal{I}^\text{gde}}
\newcommand{\Iref}{\mathcal{I}^\text{ref}}
\newcommand{\Vs}{\mathcal{V}}
\newcommand{\Itrain}{\mathcal{I}^\text{train}}
\newcommand{\Icand}{\mathcal{I}^\text{trans}}
\newcommand{\Irend}{\mathcal{I}^\text{rend}}
\newcommand{\Drend}{\mathcal{D}^\text{rend}}
\newcommand{\Mrend}{\mathcal{M}^\text{rend}}
\newcommand{\Zrend}{\mathcal{Z}^\text{rend}}
\newcommand{\F}{\mathcal{F}}
\newcommand{\fg}{\mathbf{f}}
\newcommand{\kfg}{\mathbf{k}}
\usepackage{amsmath}
\DeclareMathOperator*{\argmax}{arg\,max}
\DeclareMathOperator*{\argmin}{arg\,min}
\definecolor{deepred}{HTML}{B1001C}
\definecolor{deepblue}{HTML}{003776}
\definecolor{chocolate}{RGB}{120,60,40} 
\definecolor{silvergray}{RGB}{100,100,100}

%% file: sec/1_intro.tex
\section{Introduction}
\label{sec:1}

In the evolving field of 3D computer vision, user-friendly 3D editing has attracted growing attention as a key research focus~\cite{chen2023gaussianeditor,haque2023instructnerf2nerf,karim2023free,dong2024vicanerf,song2023efficient,chen2024dge,wang2024view,lee2024editsplat,dihlmann2024signerf,lee2023ice,zhang20243ditscene,fang2023gaussianeditor}.
Among recent advances, 3D Gaussian Splatting (3DGS)~\cite{kerbl20233d} has emerged as a groundbreaking 3D representation, offering an explicit and efficient structure that supports local manipulation and rendering in real time.
Building up the 3DGS representation, we focus on a practical and intuitive form of user interaction---personalizing a 3DGS scene using only a single-view reference image---by editing a user-specified region to match the reference appearance.
To make it clear, as illustrated in Figure~\ref{fig:1}, given a reference image depicting a unique brown \textbf{\textit{\textcolor{chocolate}{panda}}}, our goal is to modify a user-specific \textbf{\textit{\textcolor{silvergray}{bear}}} region in the scene to a \textbf{\textit{\textcolor{chocolate}{panda}}} that aligns to the reference appearance.
This task enables intuitive 3D customization from a single image, supporting applications such as personalized avatars in virtual reality and assets stylization in interactive environments.

With the advent of large-scale pre-trained 2D diffusion models~\cite{Rombach_2022_CVPR,podell2023sdxl,peebles2023scalable}, recent 3DGS editing methods~\cite{chen2024dge,wang2024view,fang2023gaussianeditor,wu2024gaussctrl} have predominately leveraged image generation models to produce pseudo-images as editing guidance that supervise the fine-tuning of 3DGS scenes.
In this paradigm, the task of image-conditioned personalization requires two key consistencies in the guidance images:
(1) \textbf{referential consistency} with the visual appearance of the reference image and
(2) \textbf{multi-view consistency} across different perspectives to prevent conflicting guidance.
However, achieving these consistencies remains a significant challenge for existing approaches~\cite{zhuang2024tip} conditioned on a single reference image.
As illustrated in Figure~\ref{fig:2}, prior methods typically adapt their image generation models directly to the single reference view, often misprojecting appearance features entangled with its geometry onto unrelated viewpoints.
This leads to distorted appearances and severe multi-view inconsistencies in the editing guidance, ultimately resulting in noticeable artifacts in the final 3D output.

We argue that the core challenge lies in the viewpoint bias introduced by the limited perspective of a single reference image, where the image model lacks sufficient information to infer appearances under novel viewpoints that are far from the reference.
As a result, the model is often biased towards the reference view, making existing methods struggle to produce consistent multi-view editing guidance.
Therefore, in this paper, we propose \textbf{Consistent Personalization for 3DGS ($\method$)}, a high-quality personalization framework that addresses the viewpoint bias by progressively propagating the reference appearance to novel perspectives.
As illustrated in Figure~\ref{fig:2}, to use a rough appearance as structural priors and establish viewpoint cues for guidance image generation, $\method$ operates in a coarse-to-fine manner with three stages:
(1) Coarse guidance generation to initialize geometry and propagate rough appearance.
(2) Iterative LoRA fine-tuning to extract and extend fine-grained reference details.
(3) View-consistent generation that leverages the coarse guidance and trained LoRA to produce the final guidance images, which are used to fine-tune and produce the 3DGS output.

In the first stage, we establish a coarse guidance that serves as a structural prior, enabling the initial propagation of reference appearance into a coarse, view-consistent 3D representation. Specifically, we employ a pre-trained image-to-3D generation model~\cite{xiang2024structured} to produce a geometry-consistent contour with a rough texture estimate, which is integrated into the target location in the scene. As shown in Figure~\ref{fig:2}, although the resulting textures are often unrealistic due to the domain gap between the real-world reference image and the CGI-style pre-training data~\cite{deitke2023objaverse,deitke2023objaversexl} of image-to-3D model, the coarse guidance reliably captures structural geometry and a rough yet view-consistent appearance.

To recover fine-grained reference appearance free from the viewpoint bias, the second stage draws inspiration from~\cite{raj2023dreambooth3d}, which shows that diffusion models adapted to a single image hold the potential to generate novel neighboring views around the reference.
Building on this insight, we propose an iterative LoRA fine-tuning strategy that gradually extracts and propagates reference appearance to novel viewpoints.
In each iteration, we translate novel-view renderings of the coarse guidance using the current model and select one well-aligned result---identified via our designed scoring mechanism based on dense feature matching~\cite{edstedt2024roma}---to augment the training set for the next round fine-tuning.

Leveraging the coarse guidance and the trained LoRA, we employ a pre-trained flow-based model~\cite{flux2024} in the last stage to generate the final guidance images.
We begin by applying rapid rectified-flow inversion~\cite{rout2024semantic} to convert renderings of the coarse guidance into noisy latents, which are passed to the Flow Transformer and serve as the starting point for generation, conditioned on the depth maps of the coarse guidance.
To further reduce the multi-view inconsistency arising from viewpoint variance, we introduce an epipolar-constrained token replacement strategy that aggregates visual features across all views based on geometric correspondences, effectively improving overall multi-view coherence.

As illustrated in Figure~\ref{fig:1}, by progressively propagating the single-view reference appearance in a coarse-to-fine manner, $\method$ effectively addresses the challenges posed by the viewpoint bias, resulting in superior visual quality in personalized 3DGS results.
Comprehensive evaluations across diverse real-world scenes demonstrate that our $\method$ successfully address the artifacts caused by limited reference perspectives and outperforms state-of-the-art methods in both qualitative and quantitative comparisons.
Based on the above, our contributions can be summarized in three aspects:

\begin{itemize}[leftmargin=*]
\item We identify the viewpoint bias caused by limited reference perspective information as the crux of referential and multi-view inconsistencies in previous single-view 3D personalization methods.
\item To mitigate the viewpoint bias, we propose a coarse-to-fine appearance propagation framework that progressively expands the single-view reference appearance to novel perspectives, generating guidance images with faithful referential consistency and strong multi-view consistency.
\item We validate $\method$ through extensive experiments on various real-world scenes, demonstrating its superior performance over previous 3DGS personalization and editing methods in both qualitative and quantitative evaluations.
\end{itemize}

%% file: sec/2_related_work.tex
\section{Related Works}

\begin{figure}[t]
\centering
\includegraphics[width=1.\linewidth]{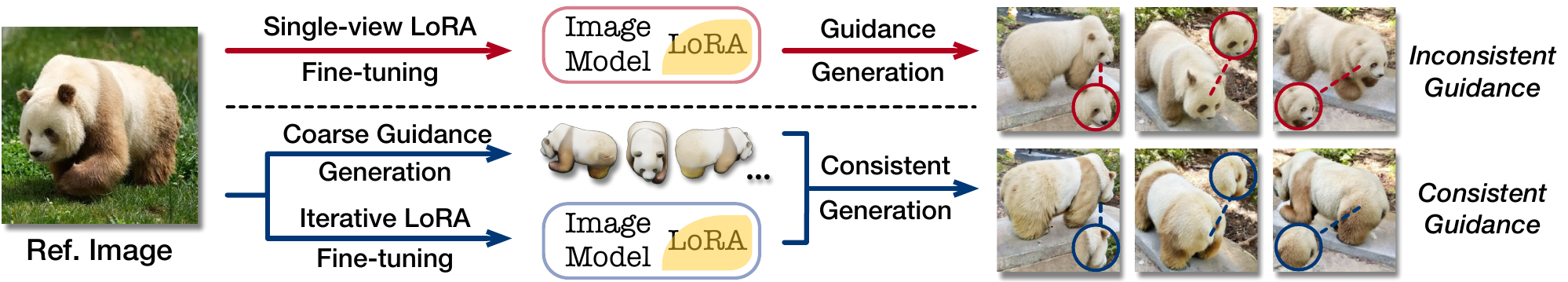}
\caption{\textbf{\textit{\color{deepred}red}}: Previous methods suffer from viewpoint bias and produce distorted editing guidance, leading to both the referential and multi-view inconsistencies. \textbf{\textit{\color{deepblue}blue}}: By progressively propagating reference to novel views, $\method$ mitigates the bias and achieves both consistencies in the guidance.}
\label{fig:2}
\end{figure}

\noindent
\textbf{Image-guided 2D Customization.}
Given a set of reference images, the task of 2D customization aims to edit a source image or generate a new image under the guidance of the reference, where customization methods~\cite{gal2022image,ruiz2023dreambooth,liu2023cones} typically optimize a special token or use LoRA-based adaptation to capture the appearance of the reference images.
Built on this strategy, early methods~\cite{gal2022image,ruiz2023dreambooth,liu2023cones,kumari2023multi,tewel2023key,zhang2024generative,choi2023custom,shi2024instantbooth,gu2023photoswap,yang2023paint,cai2024decoupled} rely on multiple reference images to construct the novel content through test-time fine-tuning (TTF).
Subsequent works~\cite{chen2023disenbooth,li2023dreamedit,sohn2023styledrop,avrahami2023break} further improve the flexibility of this paradigm by training with a single reference image.
Recently, leveraging large-scale image datasets~\cite{schuhmann2022laion,pexels,kakaobrain2022coyo-700m,changpinyo2021conceptual,sharma2018conceptual,lin2014microsoft}, a line of work~\cite{li2023blip,chen2024zero,yuan2023customnet,chen2024anydoor,li2024photomaker,chen2024dreamidentity,peng2024portraitbooth} has adopted pre-trained adaptation (PTA), which trains on large-scale paired data and bypasses fine-tuning during inference.
While our iterative LoRA fine-tuning strategy builds on the test-time fine-tuning paradigm, the task of 3D personalization presents additional challenges beyond those in 2D customization, notably the need for multi-view consistency and mitigating the viewpoint bias.

\noindent
\textbf{Consistent 3D Field Editing.}
Early approaches for consistent 3D editing~\cite{kamata2023instruct,yu2023edit,zhou2023repaint,park2023ed,khalid2023latenteditor,bao2023sine,yi2024diffusion,zhuang2023dreameditor,hertz2023delta,koo2023posterior} predominantly rely on NeRF~\cite{mildenhall2021nerf} representations optimized via Score Distillation Sampling (SDS)-based techniques~\cite{poole2022dreamfusion}.
Subsequent works~\cite{chen2023gaussianeditor,haque2023instructnerf2nerf,dong2024vicanerf} employ image-guided 3D editing by leveraging pre-trained 2D diffusion models to generate multi-view guidance images. 
Pioneered by~\cite{chen2023gaussianeditor,fang2023gaussianeditor}, recent methods integrate Gaussian Splatting~\cite{kerbl20233d} into 3D field editing due to its superior efficiency and controllability.
More recently, a line of research~\cite{chen2024dge,wang2024view,lee2024editsplat,wu2024gaussctrl} has aimed to explicitly ensure multi-view consistency in the guidance images. VcEdit~\cite{wang2024view} introduces latent and attention map aggregation, while GaussCtrl~\cite{wu2024gaussctrl} and DGE~\cite{chen2024dge} utilize cross-view extensive attention to harmonize the variations across views.
However, all these methods are limited to simple text prompts condition and lack the ability of customized editing.
The most relevant work with ours is TIP-Editor~\cite{zhuang2024tip}, which combines LoRA and SDS to distill the reference content into 3D scene. However, it fails to consistently expand the reference appearance across views, often exhibits visual artifacts in the 3D outputs due to viewpoint bias.

%% file: sec/3_method.tex
\section{Methodology}

\begin{figure}[t]
\centering
\includegraphics[width=1.\linewidth]{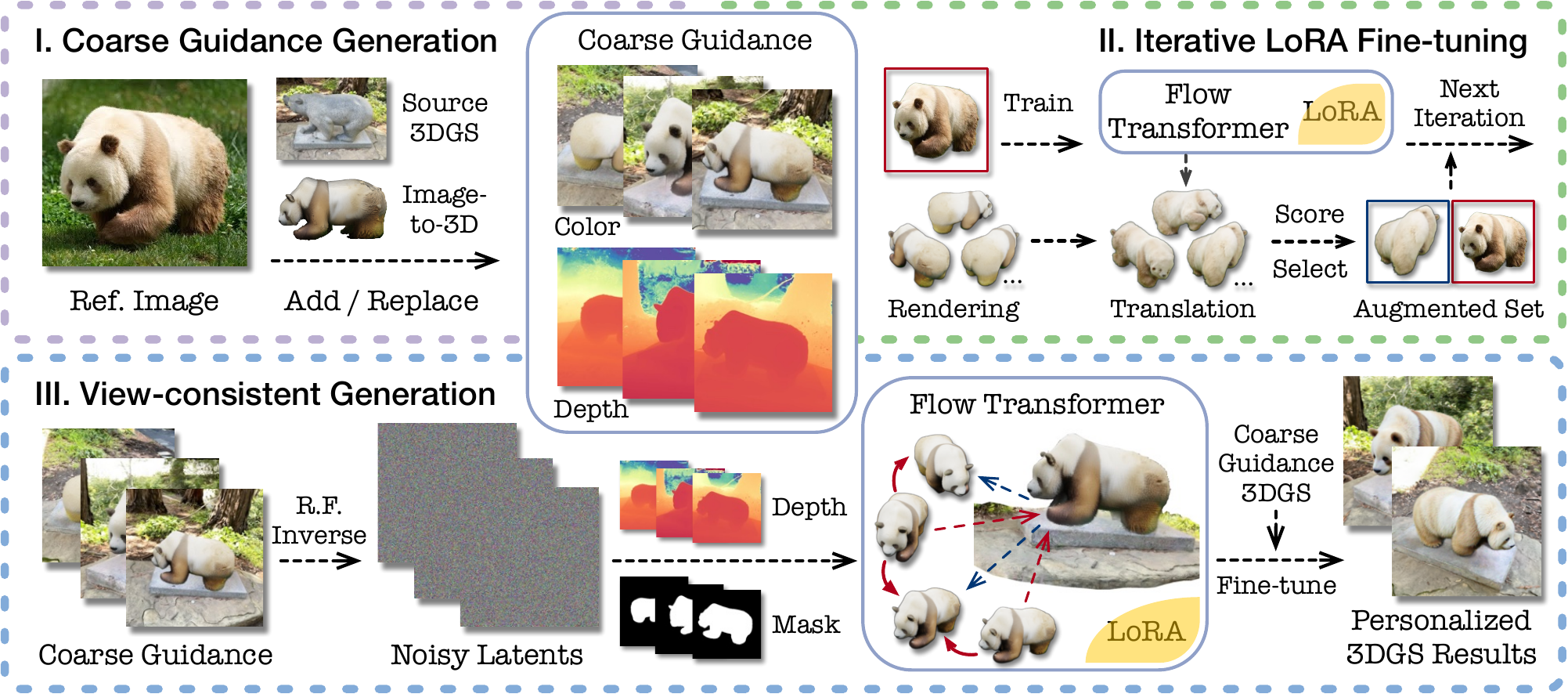}
\caption{The pipeline of our \textbf{$\method$} includes three stages: coarse guidance generation via a pre-trained image-to-3D model; iterative LoRA fine-tuning to extract and propagate detailed reference appearance; and view-consistent generation of guidance images to produce final 3DGS outputs.}
\label{fig:3}
\end{figure}

In this section, we present the \textbf{CP-GS} personalization framework from a single-view reference image (Sec.~\ref{sec:3_1}), with the overall pipeline illustrated in Figure~\ref{fig:3}.
We first employ a pre-trained image-to-3D model to construct a coarse guidance with rough yet view-consistent reference appearance, serving as the initial step of our propagation (Sec.~\ref{sec:3_2}).
To further extract and propagate fine-grained reference appearance, we then introduce an iterative LoRA fine-tuning strategy that progressively expands the training views through image translation and selective augmentation (Sec.~\ref{sec:3_3}).
Finally, we combine the coarse guidance and the trained LoRA within a pre-trained Flow model to generate multi-view consistent, reference-aligned guidance images, resulting in the final 3DGS output (Sec.~\ref{sec:3_4}).

\subsection{Problem Definition}
\label{sec:3_1}

Given a source 3DGS scene $\gsrc$ and a reference image $\Iref$, the goal is to edit a user-specific region in $\gsrc$ to the personalized $\gedt$ that align with $\Iref$.
To achieve this, we adopt an image-guided paradigm that generates a set of multi-view personalized guidance images $\Iedt$ to supervise the transformation of $\gsrc$ into the output $\gedt$.
We define an editing loss for each view by combining a mean absolute error $\mathcal{L}_\text{MAE}$ and a perceptual loss $\mathcal{L}_\text{LPIPS}$ between the real-time rendering and corresponding guidance image. The final 3DGS model $\gedt$ is optimized by minimizing the total loss across all views $\Vs$:
\begin{equation}
\gedt = \: \underset{\G}{\mathrm{argmin}} \: \sum_{v \in \Vs}{ (\lambda_1\mathcal{\mathcal{L}_\text{MAE}}(\mathcal{R}(\G, v),\: \Iedt) + \lambda_2\mathcal{\mathcal{L}_\text{LPIPS}}(\mathcal{R}(\G, v),\: \Iedt) )},
\end{equation}
where $\mathcal{R}$ denotes the rendering function~\cite{kerbl20233d}.
This paradigm requires the multi-view guidance images $\Iedt$ to satisfy two key properties: multi-view consistency across $\Vs$ to prevent optimization conflicts, and referential appearance consistency to the $\Iref$ to fulfill the personalization objective. Our $\method$ is designed to explicitly ensure both consistency to achieve high-quality 3D personalization.

\subsection{Coarse Guidance Generation}
\label{sec:3_2}

As noted in Sec.~\ref{sec:1}, the limited perspective of a single reference image fails to provide sufficient geometric and coherent appearance information for constructing a consistent multi-view representation.
Therefore, in the first stage, we leverage an off-the-shelf image-to-3D generation model TRELLIS~\cite{xiang2024structured}, pre-trained on large-scale CGI-style 3D datasets~\cite{deitke2023objaverse, deitke2023objaversexl}, to produce a coarse guidance scene that expands the reference into a rough yet multi-view consistent representation.
As illustrated in the \textit{top-left} of Figure~\ref{fig:3}, the reference image is fed into the pre-trained TRELLIS to generate the corresponding 3D rough asset, which is then integrated into the source scene to replace or augment the user-specific target region.
We provide two integration modes: (1) Adding new content – the user provides a 3D bounding box specifying the object’s position and scale;
(2) Replacing existing content – the target bounding box is extracted from the existing content via PCA~\cite{abdi2010principal}, and the generated asset is fitted accordingly.
This coarse guidance provides a plausible 3D geometry and establishing a rough yet view-consistent appearance that serves as a structural prior for subsequent stages.

\begin{wrapfigure}{r}{0.5\textwidth}
\centering
\vspace{-13pt}
\includegraphics[width=1.\linewidth]{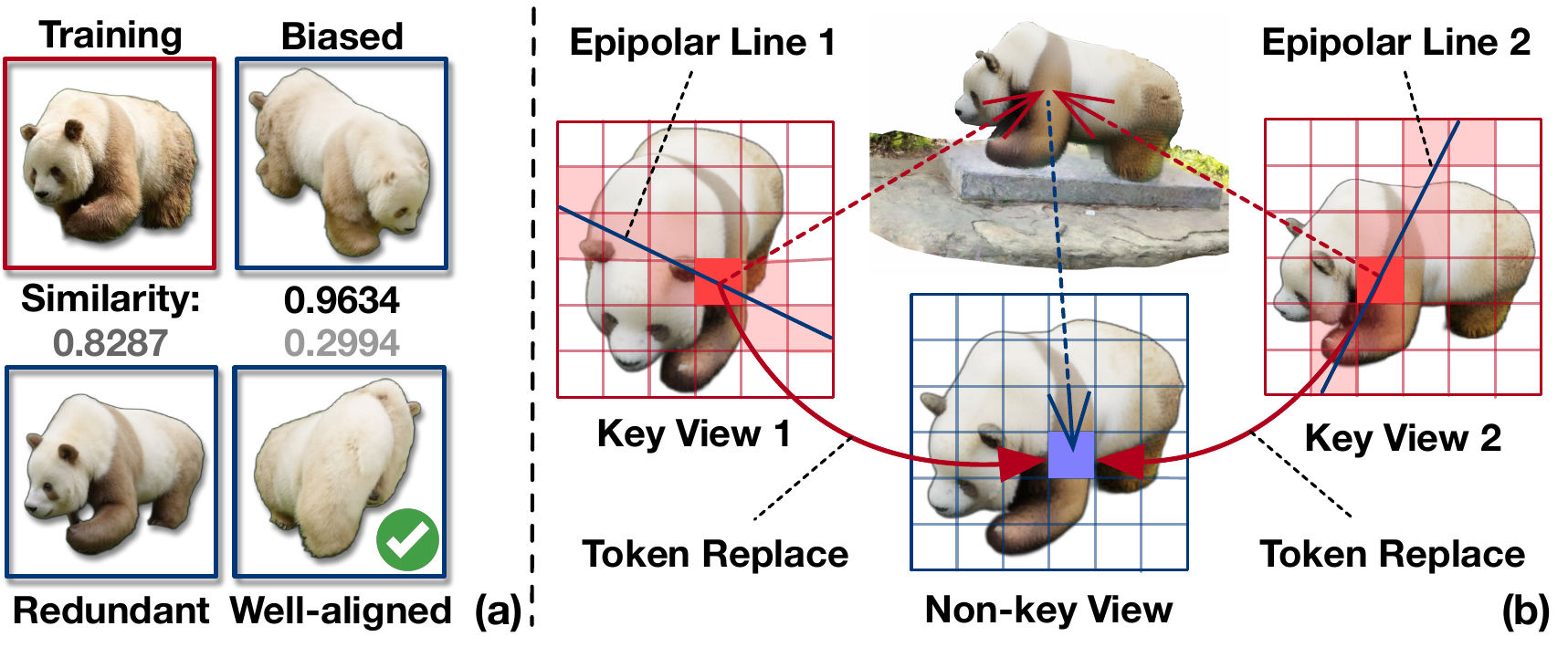}
\caption{\textbf{(a)} Visualization of the translated results and the corresponding similarities under our scoring mechanism. \textbf{(b)} Illustration of the proposed epipolar-constrained token replacement strategy.}
\label{fig:4}
\vspace{-4pt}
\end{wrapfigure}

\subsection{Iterative LoRA Fine-tuning}
\label{sec:3_3}

Due to the inevitable domain gap between the real-world reference image and the CGI-style datasets~\cite{deitke2023objaverse,deitke2023objaversexl} used to train the image-to-3D model, the generated assets in coarse guidance often exhibit unrealistic and rough appearance that lacks referential consistency.
Under the single image setting, we observe that image generation model also tends to overfit to the reference perspective, resulting in a strong viewpoint bias. 
To address these issues, we adopt an iterative LoRA~\cite{hu2022lora} fine-tuning strategy that retrieves a fine-grained appearance from the reference and progressively propagates it to novel views.

Specifically, we initialize the LoRA training set with the given single-view image, conducting the first iteration of fine-tuning using a prompt containing a special token to encode the reference characteristics. Inspired by DreamBooth3D~\cite{raj2023dreambooth3d}, which demonstrates that image generation models~\cite{Rombach_2022_CVPR} adapted to a single image can synthesize novel views of the reference subject within a limited range around the training perspective, we render the coarse guidance from multiple viewpoints and apply the fine-tuned model to translate their appearance toward the reference target using the same prompt.
Subsequently, we select one well-aligned translated image using a task-specific scoring mechanism and append it to the training set to augment the next round of fine-tuning.

We notice that designing such scoring mechanism is non-trivial, as it must avoid viewpoint-biased translations and redundant views that are already well covered by the training set, ensuring that each selection contributes meaningfully to appearance propagation.
Notably, as shown in Figure~\ref{fig:4}(a), both types of undesirable cases tend to exhibit high similarity to the training images:
(1) redundant views, which are close to the training perspective, naturally share similar appearance; and
(2) biased translations, which often inherit excessive training-view features due to overfitting, also tend to exhibit higher similarity to training images than the well-aligned novel-view results.
Therefore, we identify the well-aligned result as the one with minimal overall similarity to the training set, measured via dense feature matching using the pre-trained RoMa model~\cite{edstedt2024roma}.
Denoting $\Itrain_t$ the training image set and $\Icand_t$ the translations at iteration $t$, our scoring and selection are formulated as:
\begin{equation}
    \Itrain_{t+1} = \Itrain_{t} \cup \{ \argmin_{i} \sum_{j} \mathbf{S}_\text{RoMa}(I^\text{trans}_i, I^\text{train}_j) \}, \:\: \text{where} \:\: I^\text{trans}_i \in \Icand_t, \:\: I^\text{train}_j \in \Itrain_t
\end{equation}
where $\mathbf{S}_\text{RoMa}(\cdot)$ denotes the similarity computed by RoMa~\cite{edstedt2024roma} model.
Leveraging the neighboring-view generation capability of the LoRA module, our iterative fine-tuning strategy effectively propagates single-view reference details to novel perspectives and alleviates the viewpoint bias, enabling the model to learn fine-grained appearance with both multi-view and reference consistency.

\begin{figure}[t]
\centering
\includegraphics[width=1.\linewidth]{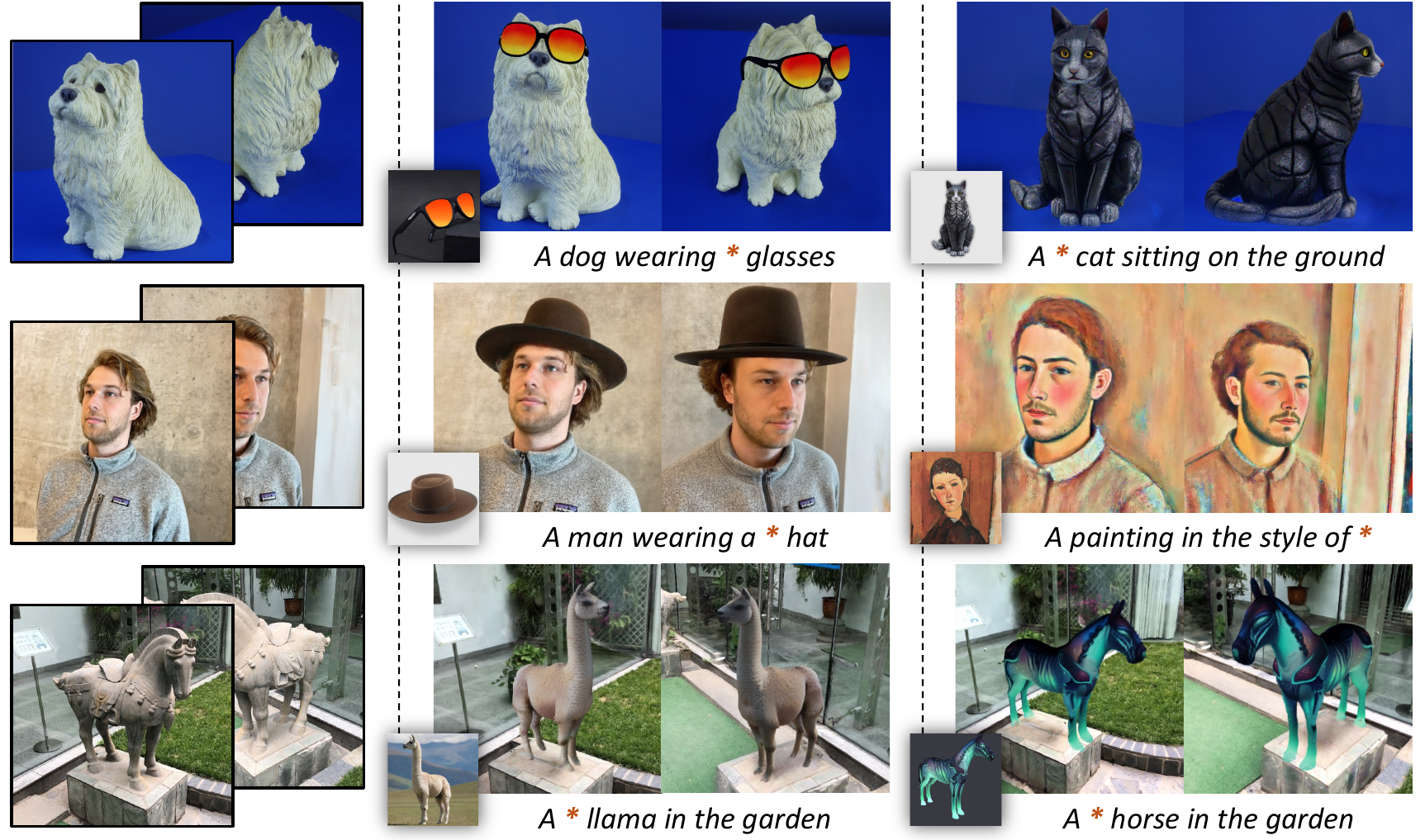}
\caption{Additional personalization result of our \textbf{$\method$}, demonstrating high-quality 3DGS scene customization that faithfully align with the reference image across various scenarios.}
\label{fig:5}
\end{figure}

\begin{figure}[t]
\centering
\includegraphics[width=1.\linewidth]{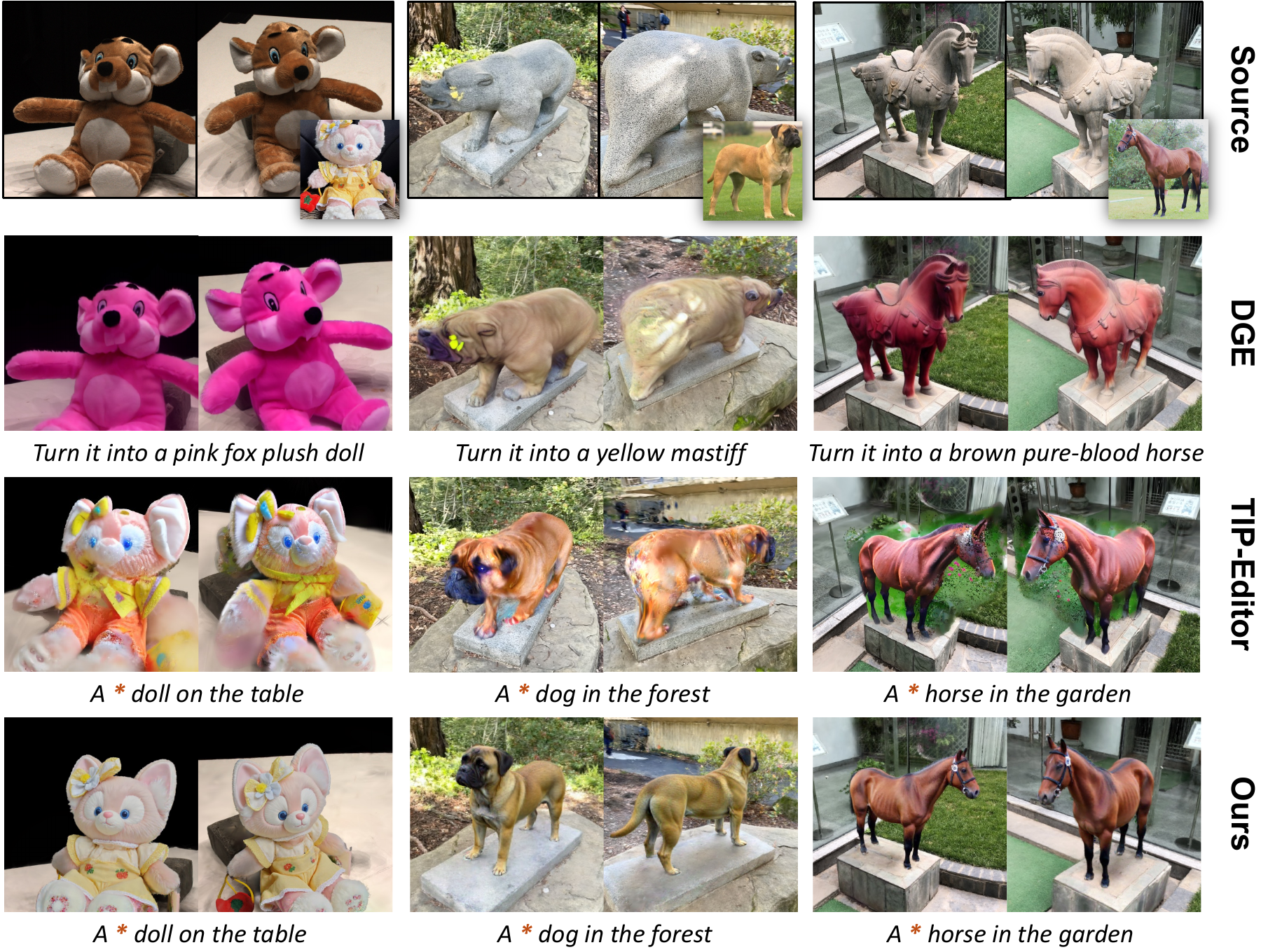}
\caption{Qualitative comparison of personalization results between our $\method$ and the existing SOTA methods~\cite{chen2024dge, zhuang2024tip}, where $\method$ outperforms with superior visual quality and reference alignment.}
\label{fig:6}
\end{figure}

\subsection{View-consistent Generation}
\label{sec:3_4}
In the final stage, we adopt a view-consistent generation strategy based on a pre-trained Flow Transformer~\cite{flux2024}, combining the coarse guidance (Sec.~\ref{sec:3_2}) and the iteratively trained LoRA module (Sec.~\ref{sec:3_3}) to produce the final consistent guidance images.
As shown in Figure~\ref{fig:3}, we begin by rendering the coarse guidance into multi-view images and converting them into noisy latents using rectified-flow inversion~\cite{rout2024semantic} to encode both the appearance and geometry. 
Serving as the starting point for subsequent generation process, these latents are then fed into the Flow Transformer along with rendered depth maps, which provide geometric cues to align appearance generation with the underlying structure and enhance multi-view consistency.

Inspired by~\cite{chen2024dge,feng2025personalize}, we introduce an epipolar-constrained token replacement mechanism, to promote multi-view consistency by unifying foreground tokens across views that correspond to the same 3D locations.
We perform token replacement in the early dual-stream blocks of the Flow Transformer, where visual tokens are explicitly maintained and can be directly modified.
During generation, we automatically select a set of key views with minimal overlap to ensure full coverage, and extract foreground pixel indices in all viewpoints using masks from the coarse guidance.
As illustrated in Figure~\ref{fig:4}(b), for each foreground token in non-key views, we compute its epipolar lines on the two nearest key views and replace it with an interpolation of the most similar tokens along those epipolar lines, weighted by the camera distance to each key view.
Given a non-key frame, the interpolated token $\fg'(\upix)$ at pixel $\upix$ used to replace the original token $\fg(\upix)$ is computed using the foreground tokens of the two nearest key frames $\kfg_i$, indexed by $i \in \{1, 2\}$. Letting $c$ denote the non-key camera and $l_{\upix \rightarrow i}$ denote the epipolar line of $\upix$ in each key view, the token $\fg’(\upix)$ is computed as:
\begin{equation}
    \fg'(\upix) = \sum_i \kfg_i(\vpix_i)\mathcal{D}(c, c_i)/\sum_{i}\mathcal{D}(c, c_i), \:\:\text{where}\:\: \vpix_i = \underset{\vpix \in l_{\upix \rightarrow i}}{\argmax} \: \langle\fg(\upix), \kfg_i(\vpix)\rangle
\end{equation}
where $\mathcal{D}(c, c_i)$ represents the camera distance from $c$ to each key view's camera $c_i$.
This mechanism effectively alleviates cross-view variance, producing guidance images with strong multi-view consistency and faithful reference alignment.
These images then supervise the 3DGS parameter updating of the coarse guidance, yielding the final personalized 3DGS result of our CP-GS framework.

%% file: sec/4_experiment.tex
\section{Experiments}

\subsection{Implementation Details}
\label{sec:4_1}
We implement our framework based on the official 3DGS codebase~\cite{kerbl20233d}, GaussianEditor~\cite{chen2023gaussianeditor}, and the LoRA training scripts from Diffusers~\cite{von-platen-etal-2022-diffusers}. We employs TRELLIS~\cite{xiang2024structured} as image-to-3D model to generate our coarse guidance.
For the Flow transformer, we adopt FLUX.1-dev~\cite{flux2024} equipped with the depth LoRA adapter.
In most cases, we use two iterations of LoRA training to propagate the appearance, which takes around 15 minutes on two NVIDIA RTX A6000 GPUs and is reusable across different source scenes.
Using the trained LoRA and coarse guidance, we generate consistent multi-view guidance images and optimize the 3DGS model with the Adam optimizer~\cite{kingma2014adam} at a learning rate of 0.001, taking around 10 minutes per scene when using the same two A6000 GPUs.

\subsection{Qualitative Evaluation}
\label{sec:4_2}

We compare our $\method$ with two state-of-the-art 3DGS editing baselines: DGE~\cite{chen2024dge}, which conditioned on text prompt, and TIP-Editor~\cite{zhuang2024tip}, the only existing method with the same single-image condition as ours.
We construct a challenging test set comprising reference images from TIP-Editor and additional internet-sourced examples with highly specialized, visually intricate appearances.
For DGE, we employ GPT-4o~\cite{hurst2024gpt} to generate concise captions (within 5 words) that describe the reference object, as longer prompts were observed to degrade its performance.
As shown in Figure~\ref{fig:6}, both baselines fail to preserve the distinctive appearance features of the reference images. Moreover, TIP-Editor exhibits severe artifacts in its personalized results, primarily due to multi-view inconsistencies in the guidance images resulting from viewpoint bias.
In contrast, our $\method$ consistently produces clean, coherent, and intricately detailed edits that faithfully align with the reference image.

This performance gap underscores the inability of the baseline methods to capture and propagate reference appearance.
In particular, DGE illustrates the shortcomings of text-conditioned 3DGS editing for personalization: lacking direct access to the reference image, it relies solely on short textual prompts that fail to capture rich visual details.
Moreover, the specialized reference images often fall outside the distribution of text-to-image models, making them difficult to represent accurately.
On the other hands, TIP-Editor lacks explicit mechanism to extend the reference appearance to novel viewpoints, resulting in strong viewpoint bias, which introduces multi-view inconsistencies in its 2D guidance, ultimately leading to visual artifacts in the 3DGS results.
By contrast, $\method$ explicitly addresses these issues through the coarse-to-fine appearance propagation, enabling high-quality 3DGS personalization that ensures both referential and multi-view consistency.
Additional results showcasing the effectiveness of our $\method$ are presented in Figure~\ref{fig:5} and the \textit{Appendix}.

\begin{figure}[t]
\centering
\includegraphics[width=1.\linewidth]{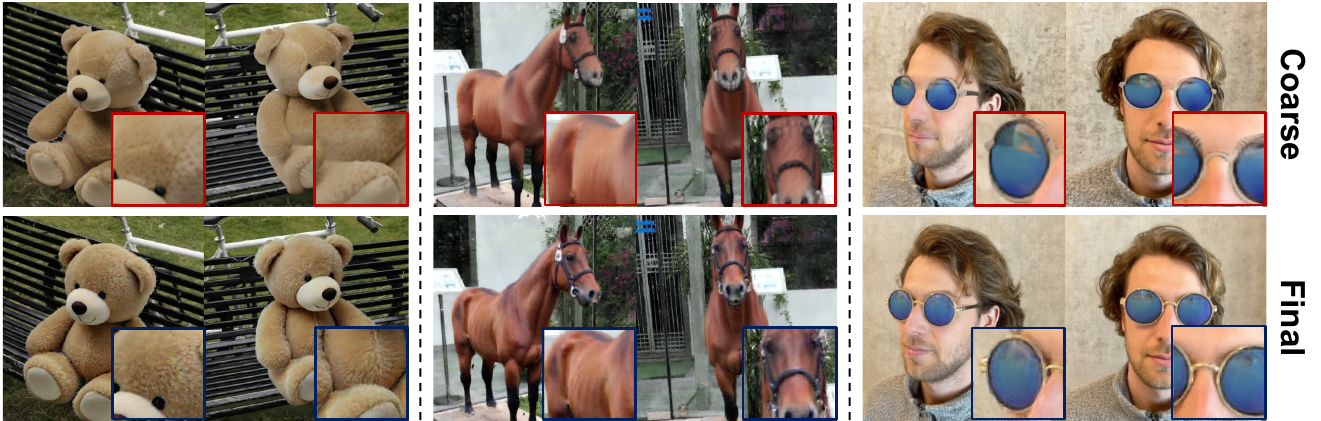}
\caption{Comparison between the coarse guidance and our final results, where our final results effectively refine the unrealistic and visually discordant appearance presented in the coarse guidance.}
\label{fig:7}
\end{figure}

\begin{table}[t]
\centering
\caption{Quantitative comparison between our $\method$ and the existing SOTA methods~\cite{chen2024dge,zhuang2024tip}, where $\method$ significantly outperform others in both visual quality and the alignment with reference image.}
\resizebox{0.95\linewidth}{!}{
\begin{tabular}{lccccc}
\toprule
Methods                          & $\text{User}_{quality}\uparrow$  & $\text{User}_{align} \uparrow$ & $\text{DINO}_{sim}\cite{oquab2023dinov2}\uparrow$ & $\text{CLIP}_{sim}\cite{radford2021learning}\uparrow$ & $\text{CLIP}_{dir}\cite{gal2022stylegan}\uparrow$ \\ \midrule
DGE~\cite{chen2024dge}           & 31.89\%  & 6.37\%   & 41.73  & 67.26   & 14.22   \\
TIP-Editor~\cite{zhuang2024tip}  & 25.46\%  & 17.28\%  & 43.88  & 70.92   & 14.46   \\
$\method$ (Ours)                 & \textbf{78.28\%} & \textbf{80.09\%} & \textbf{50.33}  & \textbf{76.78}   & \textbf{18.03} \\ \bottomrule
\end{tabular}
}
\label{tab:1}
\end{table}

\begin{figure}[t]
\centering
\includegraphics[width=1.\linewidth]{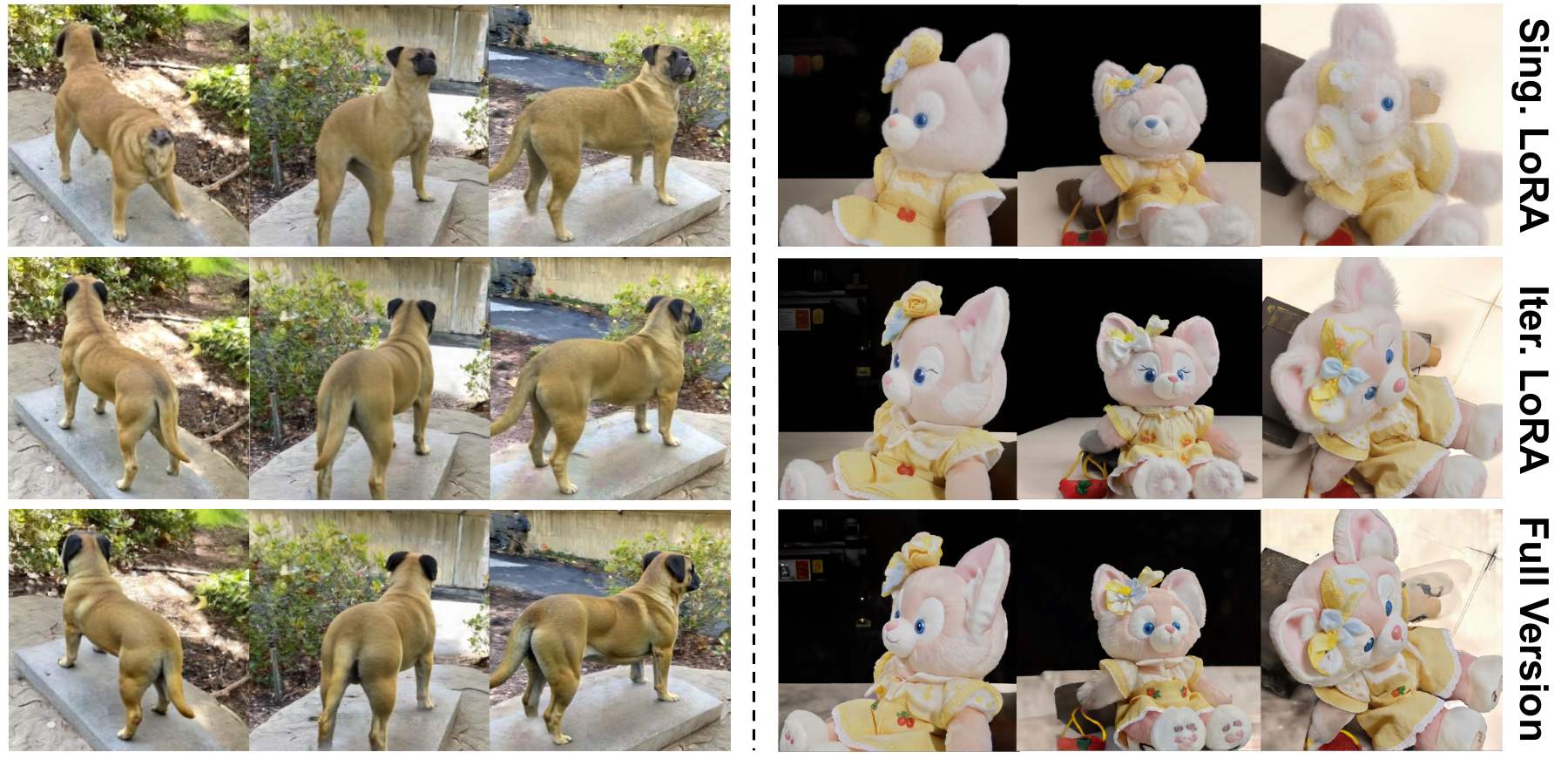}
\caption{Ablation comparison on the guidance images produced by three specific configurations: training the LoRA module only on the single-view reference image (\textit{Sing. LoRA}), using our iterative LoRA fine-tuning yet excluding the constrained token replacement (\textit{Iter. LoRA}), and our \textit{Full Version}.}
\label{fig:8}
\end{figure}

\subsection{Quantitative Evaluation}
\label{sec:4_3}

In Table~\ref{tab:1}, we present a quantitative evaluation comparing our $\method$ with the two baselines~\cite{chen2024dge,zhuang2024tip} on over 20 samples collected in the same manner as the qualitative experiments.
We first conduct a user study to assess the proportion of results deemed satisfactory by users in aspects of visual quality and the reference alignment, with further details provided in the \textit{Appendix}.
Besides, we follow existing setting~\cite{zhuang2024tip} to report CLIP~\cite{radford2021learning} and DINO~\cite{oquab2023dinov2} image-to-image similarity metrics that quantify the alignment between edited outputs and the reference image by computing visual feature similarity.
In addition, we adopt the CLIP directional similarity~\cite{gal2022stylegan} to measure the semantic alignment between the text prompt and the semantic shift from the source to edited results.
As shown in Table~\ref{tab:1}, our $\method$ consistently outperforms all baselines across both perceptual metrics and user study evaluations.
This performance gap stems from two main limitations in these baselines: (1) DGE lacks direct access to the reference image and relies solely on short text prompts, which fail to capture fine-grained appearance details; (2) TIP-Editor fails to propagate the reference appearance to novel views, resulting in strong viewpoint bias that introduces multi-view conflicts in the editing guidance and ultimately leads to visual artifacts and poor reference alignment.
These results highlight the superior performance of our $\method$, underscoring the critical role of capturing and expanding reference appearance across novel viewpoints in tackling single-view conditioned 3DGS personalization.

\subsection{Ablation Study}
\label{sec:4_4}

To analysis the contribution of each component, in this section, we compare the coarse guidance with our final results and conduct ablation studies on the iterative LoRA fine-tuning strategy and epipolar-constrained token replacement. Additional quantitative ablation study is provided in \textit{Appendix}.

\noindent
\textbf{Coarse Guidance vs. Final Results.}
In the first stage of our method, a coarse asset is generated by the image-to-3D model (Sec.~\ref{sec:3_2}) and integrated into the source scene to produce the coarse guidance. A natural question arises: \textit{can this coarse guidance suffice as the final edit?} To assess the necessity of our subsequent stages, we compare the coarse guidance with our final personalization results in Figure~\ref{fig:7}. 
This comparison reveals two major limitations of the coarse guidance:
(1) The domain gap between real-world reference images and the CGI-style training data of the 3D generation model~\cite{xiang2024structured} results in overly smooth and grid-like unrealistic textures that fail to reflect the photorealic appearance of the reference.
(2) Direct insertion of the generated asset leads to poor contextual blending, where the inserted object often appears visually detached from the 3DGS scene (e.g., the sunglasses example), especially around boundaries.
In contrast, our final results exhibit rich, realistic textures that closely resemble the reference image and blend seamlessly with the source 3DGS scene, demonstrating the effectiveness and necessity of the subsequent refinement stages.

\noindent
\textbf{Iterative LoRA Fine-tuning.}
We compare two configurations to assess the contribution of our iterative LoRA fine-tuning: using only the single-view reference image for fine-tuning (\textit{Sing. LoRA}) versus applying our iterative expansion strategy (\textit{Iter. LoRA}) introduced in Sec.~\ref{sec:3_3}.
The \textit{top 2 rows} in Figure~\ref{fig:8} shows the resulting guidance images from these variants.
In the presented viewpoints that deviate from the reference, \textit{Sing. LoRA} misprojects the appearance entangled with reference-view geometry to unrelated views, resulting in noticeable distortion and multi-view inconsistency.
In contrast, \textit{Iter. LoRA} significantly alleviates this distortion, generating appearance correctly adapted to these novel perspectives.
This highlights the importance of progressively expanding the reference coverage in mitigating the viewpoint bias and producing multi-view consistent guidance images.

\noindent
\textbf{Constrained Token Replacement.}
To further reduce the cross-view variance during generation, we adopt an epipolar-constrained token replacement strategy during the generation of final guidance images (Sec.~\ref{sec:3_4}).
The \textit{bottom 2 rows} in Figure~\ref{fig:8} compares the \textit{Full Version} that includes this mechanism and the \textit{Iter. LoRA} variant where it is disabled.
Although \textit{Iter. LoRA} successfully expands the reference appearance to novel viewpoints and resolves major distortions, it still suffers from subtle multi-view inconsistencies in visual details (e.g., the mouth orientation of the \textit{dog}, and the \textit{doll}’s eyelashes).
In contrast, our \textit{Full Version} leverages 3D-aware token replacement guided by epipolar constraints and eliminate such inconsistencies, producing guidance images with improved multi-view appearance consistency that enhance the coherence and visual fidelity of the final 3DGS results.

%% file: sec/5_conclusion.tex
\section{Conclusion}
\label{sec:5}

In this paper, we presented $\method$, a novel framework for consistent and personalized 3D scene editing from a single-view reference image. 
To address the visual artifacts in existing image-conditioned methods caused by viewpoint bias and limited reference perspective, $\method$ introduces a coarse-to-fine reference propagation framework that integrates coarse guidance generation, iterative LoRA fine-tuning, and a view-consistent generation stage leveraging geometric cues and epipolar-constrained token replacement.
These components enable the generation of guidance images with strong multi-view consistency and faithful referential consistency, producing high-quality 3DGS results.
Extensive experiments across real-world scenes demonstrate that $\method$ markedly outperforms existing methods in both visual quality and reference alignment, enabling high-quality 3DGS personalization for real-world applications.
In the future, we aim to improve $\method$’s efficiency by distilling it into a single-pass pipeline and enhancing robustness to occluded reference images.

%% file: supp_sec/1_more_imp.tex
\section{More Implementation Details}
\label{supp_sec:imple}

\subsection{Step-by-step Demonstration of $\method$}
In our main paper, we introduce our $\method$ personalization framework that progressively propagate the single-view reference image appearance to novel perspectives. In Algorithm~\ref{algo:1}, we provide a step-by-step demonstration of our entire pipeline for content adding or replacement, including the coarse guidance generation, the iterative LoRA fine-tuning, the view-consistent generation of guidance images, and the final 3DGS optimization that produces our personalized 3DGS results.

\begin{algorithm}[h]
\scriptsize
\caption{Step-by-Step Pipeline of $\method$}
\begin{algorithmic}[1]

\State \textbf{Input:} Source 3DGS scene $\gsrc$ with view set $\Vs$, single-view reference image $\Iref$, and user-specific target region.

\State Forward $\Iref$ into the pre-trained image-to-3D~\cite{xiang2024structured} model, generating a coarse 3D asset $\gcoar$.

\State Integrate $\gcoar$ into the target region of $\gsrc$, producing coarse guidance scene $\G$.

\State Render $\G$ from $\Vs$ into multi-view images $\Irend$, depth maps $\Drend$, and masks $\Mrend$ indicating $\gcoar \in \G$.

\State Initialize the image model $\F$ by a pre-trained Flux~\cite{flux2024} model and LoRA module.

\State Initialize the training image set for LoRA fine-tuning as $\Itrain_1 = \{\Iref\}$.

\State \textbf{for} $t = 1, 2, ... T$ \textbf{do} ($T=2$ as default):
    
    \State $\;\;\;\;\:\:\:$ Fine-tune the LoRA module in $\F$ using $\Itrain_t$, where the background of $\Itrain_t$ is excluded.

    \State $\;\;\;\;\:\:\:$ Translate the multi-view $\Irend$ into $\Icand$ by $\Icand = \F(\Irend)$.

    \State $\;\;\;\;\;$ For $I^\text{trans}_i \in \Icand_t$, compute similarity $\mathbf{S}_i = \underset{j}{\sum} \: \mathbf{S}_\text{RoMa}(I^\text{trans}_i, I^\text{train}_j)$ towards $\Itrain_t = \{I^\text{train}_j\}$.

    \State $\;\;\;\;\;$ Extend $\Itrain_{t+1} = \Itrain_{t} \cup \{ \underset{i}{\argmin} \: \mathbf{S}_i \}$ by the one translation with minimal $\mathbf{S}_i$.

\State Select multiple key views $\kfg$ from $\Irend$ by their perspective coverage.

\State For the rest non-key view $\fg$, find the $2$ nearest key views $\kfg_i, i\in \{1, 2\}$ by camera distance $\mathcal{D}(c, c_i)$.

\State Compute the epipolar lines $\l_{\upix \rightarrow i}$ in $\kfg_i$ corresponding to each pixel $\upix$ in non-key views $\fg$.

\State Convert $\Irend$ to noisy latents $\Zrend$ using Rectified Flow Inversion~\cite{rout2024semantic}.

\State Forward $\Zrend$, $\Drend$, and $\Mrend$ into $\F$, generating guidance images $\Iedt = \F(\Zrend, \Drend, \Mrend)$.
    
    \State $\;\;\;\;\;$ In dual-stream blocks of $\F$, for pixel $\upix \in \Mrend$ \textbf{do}: 

    \State $\;\;\;\;\;\;\;\;\;\;$ $\fg(\upix) \leftarrow \sum_i \kfg_i(\vpix_i)\mathcal{D}(c, c_i)/\sum_{i}\mathcal{D}(c, c_i), \:\:\text{where}\:\: \vpix_i = \underset{\vpix \in l_{\upix \rightarrow i}}{\argmax} \: \langle\fg(\upix), \kfg_i(\vpix)\rangle$.

\State Optimize the coarse guidance: $\gedt = \: \underset{\G}{\mathrm{argmin}} \: \sum_{v \in \Vs}{ (\lambda_1\mathcal{\mathcal{L}_\text{MAE}}(\mathcal{R}(\G, v),\: \Iedt) + \lambda_2\mathcal{\mathcal{L}_\text{LPIPS}}(\mathcal{R}(\G, v),\: \Iedt) )}$.

\State \textbf{Output:} personalized 3DGS scene $\gedt$.

\end{algorithmic}
\label{algo:1}
\end{algorithm}

\subsection{Coarse Asset Integration}
\label{supp_sec:integrate}

In our main paper, we introduce two integration strategies for generating coarse guidance and support three types of personalization from a single reference image.

For \textbf{adding} a new object to the scene (e.g., the man wearing sunglasses), we allow the user to specify a 3D bounding box indicating the desired location and scale of the inserted object. The coarse asset generated by the image-to-3D model~\cite{xiang2024structured} is then placed accordingly.
For \textbf{style transfer}, coarse guidance is not required since there are no significant geometric changes between the source and personalization results. Instead, we directly extract the reference appearance from the input image and perform the view-consistent generation stage to produce the final guidance images.

For \textbf{replacing} an existing object (e.g., replacing the \textit{bear} with a \textit{dog}) in the source scene, we first remove the original content by partially adopting the deletion pipeline from GaussianEditor~\cite{chen2023gaussianeditor}. Specifically, we use SegmentAnything~\cite{kirillov2023segany} to segment the foreground from multi-view renderings, then project the segmented regions into 3D space to identify the corresponding Gaussians. These Gaussians are subsequently pruned from the scene. Then we use SDXL~\cite{podell2023sdxl} to inpaint and fix the resulting hole caused by the deletion, following the repairing pipeline of GaussianEditor~\cite{chen2023gaussianeditor}.

To insert the new object, we provide a PCA-based~\cite{abdi2010principal} bounding box extraction tool to assist in estimating the bounding box of the original object (e.g., the \textit{bear}). 
We first determine the up direction from the “\textit{ground}” Gaussians extracted using the same segmentation method, identifying it as the principal component axis corresponding to the smallest eigenvalue in PCA.
The forward direction is then derived from the removed original object by finding the PCA axis with the largest eigenvalue within the plane orthogonal to the up direction. 
The right direction is computed as the cross product of the forward and up vectors.
We project all Gaussians of the original object onto these axes and compute the min–max range along each axis to define the 3D bounding box, discarding outliers by retaining the central 98\% of the data along each dimension.
This PCA-based bounding box extraction utility works well in most cases where the inserted object is expected to follow the original orientation and is placed upright. 
For custom insertions, users can manually adjust the estimated bounding box to indicate the desired position and alignment.

\subsection{Extension of LoRA Training Set}

For the translation process in our iterative LoRA~\cite{hu2022lora} fine-tuning, we use the same positive prompt as in the LoRA training while adopting a unified negative prompt: “\textit{ugly, deformed, disfigured, poor details, bad anatomy, cartoon, CGI, unrealistic}”, to suppress undesired artifacts and improve translation quality.
In the selective augmentation of the LoRA training set, to compute the RoMa~\cite{edstedt2024roma} similarity score between each translated image and each training image, we compute the average confidence of fixed 10,000 matching points predicted by the RoMa model, following the default configuration defined in their official implementation.

\subsection{Details of User Study}

\begin{figure}[t]
\centering
\includegraphics[width=1.\linewidth]{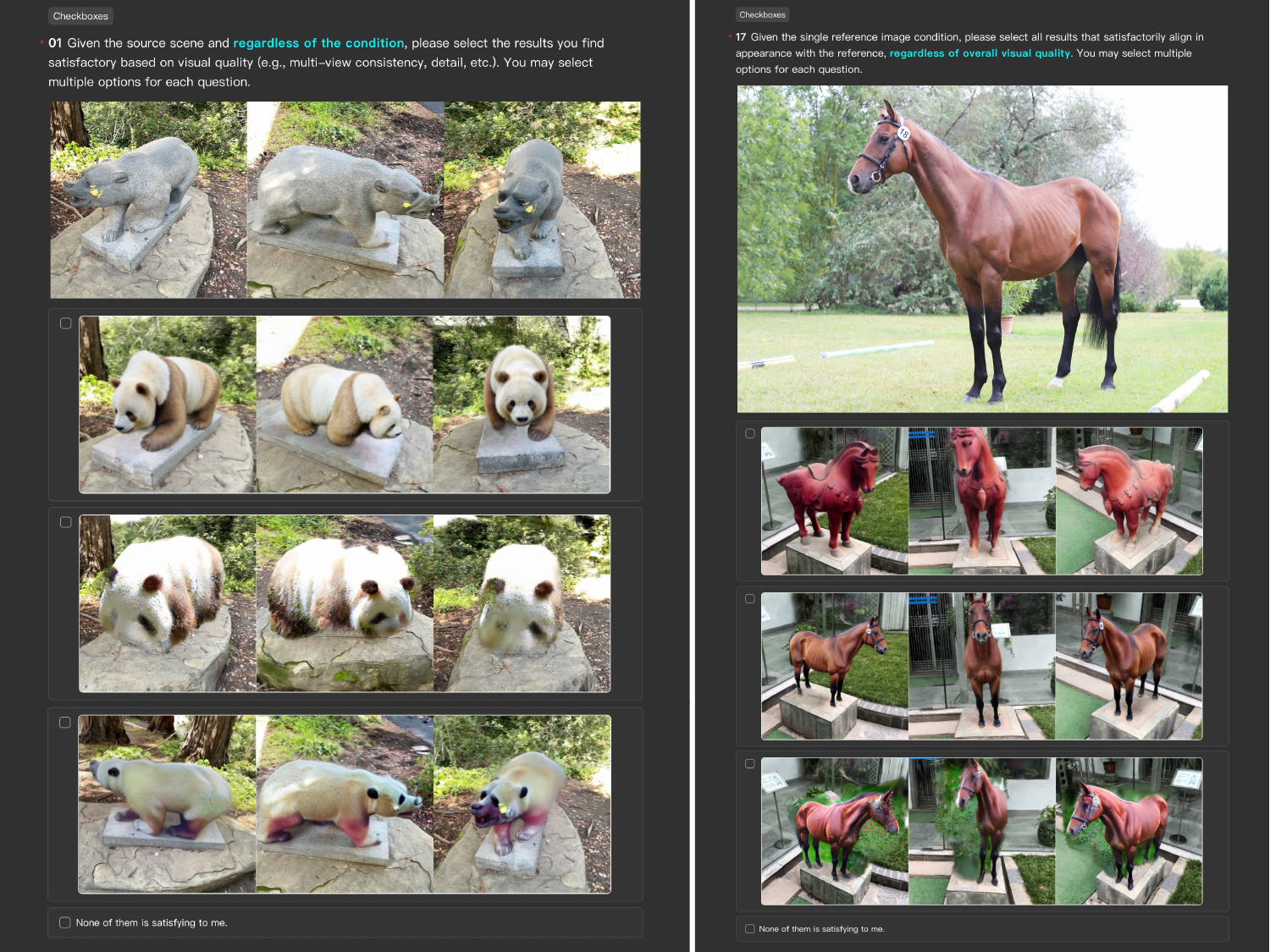}
\caption{The interface of our user study includes two types of questions: 
(1) Given the rendered source scene and the 3DGS results from each compared method, users are asked whether they are satisfied with the visual quality of each result. 
(2) Given the reference image and the renderings of each 3DGS result, users are asked to provide their satisfactory on the referential alignment.}
\label{supp_fig:userstudy}
\end{figure}

In this section, we provide additional details regarding our User Study.
As illustrated in Figure~\ref{supp_fig:userstudy}, the study consists of two types of questions aimed at evaluating two key aspects of each personalized result: visual quality and referential alignment with the input image.
To ensure fairness and avoid bias, the method names are anonymized and the order of answer choices is randomized for each question. Participants are asked to indicate whether they are satisfied with each presented result. For every method, we aggregate the responses and compute the average satisfaction rate across all samples and participants.
The study comprises 20 questions for each question type, spanning 20 representative samples and 60 personalized results from three compared methods. In total, we collected responses from 33 participants with diverse academic and professional backgrounds.
This carefully designed questionnaire, combined with a broad participant base, ensures the reliability of the results.

%% file: supp_sec/2_ext_exp.tex
\section{Extensive Experiments}
\label{supp_sec:exp}

\subsection{More Qualitative Results}
\label{supp_sec:more_quali}

As a supplement to our main paper, we present additional personalization results produced by $\method$ in Figure~\ref{fig:9}, where $\method$ effectively mitigates viewpoint bias and delivers high-quality personalization outcomes for each sample. These results further demonstrate the adaptability of $\method$ across diverse and complex scenarios with varied reference images.

\begin{table}[h]
\centering
\caption{Quantitative ablation study of our $\method$ across four configurations: the \textit{Coarse Guidance} directly integrating the generated asset, the \textit{Single-view LoRA} trained on the single reference image, the \textit{Iterative LoRA} with iterative LoRA fine-tuning, and the \textit{Full Version} further incorporating the epipolar-constrain token replacement mechanism.}
\resizebox{0.85\linewidth}{!}{
\begin{tabular}{lcccc}
\toprule
Methods                          & Coarse Guidance  & Single-view LoRA & Iterative LoRA & Full Version \\ \midrule
$\text{DINO}_{sim}\cite{oquab2023dinov2}\uparrow$      & 46.40    & 44.07    & 49.73  & \textbf{50.33}   \\
$\text{CLIP}_{sim}\cite{radford2021learning}\uparrow$  & 73.99    & 71.01    & 75.70  & \textbf{76.78}   \\
$\text{CLIP}_{dir}\cite{gal2022stylegan}\uparrow$      & 16.77    & 14.85    & 17.36  & \textbf{18.03}   \\ \bottomrule
\end{tabular}
}
\label{supp_tab:ablation}
\end{table}

\subsection{Ablated Quantitative Evaluations}
\label{supp_sec:abla_quanti}

In our main paper, we conducted a qualitative ablation study on each component of $\method$, including the coarse guidance, iterative LoRA fine-tuning, and epipolar-constrained token replacement. Here, we extend this analysis with a quantitative evaluation using the same test samples. Table~\ref{supp_tab:ablation} compares four configurations:
(1) \textit{Coarse Guidance}, which directly integrates the image-to-3D asset generated by TRELLIS~\cite{xiang2024structured};
(2) \textit{Single-view LoRA}, which trains the LoRA module solely on the single reference image for appearance refinement;
(3) \textit{Iterative LoRA}, which applies our iterative LoRA fine-tuning but excludes the epipolar-constrained token replacement; and
(4) the \textit{Full Version} $\method$, which incorporates all components.
We adopt the same evaluation metrics as in the main paper: CLIP~\cite{radford2021learning} and DINO~\cite{oquab2023dinov2} image-to-image similarity, as well as CLIP directional similarity~\cite{gal2022stylegan}. Lower values across these metrics indicate better alignment between the personalization outputs and the reference image, reflecting stronger referential consistency that fulfills the objective of the task.

As shown in Table~\ref{supp_tab:ablation}, the comparison of quantitative results align with our qualitative findings, confirming that the \textit{Full Version} of $\method$ outperforms all ablated variants. Notably, the iterative LoRA fine-tuning contributes most significantly to performance gains by effectively propagating fine-grained reference appearance to novel views (see \textit{Iterative LoRA} vs. \textit{Single-view LoRA}). The addition of epipolar-constrained token replacement further enhances multi-view consistency and improves the visual quality of the final 3DGS output (see \textit{Full Version} vs. \textit{Iterative LoRA}).
An interesting observation is that the \textit{Single-view LoRA} performs noticeably worse than the \textit{Coarse Guidance}, suggesting that naive refinement using a viewpoint-biased image generation model can degrade the quality of the coarse 3DGS asset generated from image-to-3D model~\cite{xiang2024structured}. In contrast, both the \textit{Iterative LoRA} and \textit{Full Version} show marked improvements over the \textit{Coarse Guidance}, demonstrating their effectiveness in addressing the viewpoint bias caused by limited reference perspective and generating consistent editing guidance.
These findings confirm the importance of progressive reference appearance propagation for achieving strong referential alignment in final 3DGS personalization results.

%% file: supp_sec/3_other_discuss.tex
\section{Other Discussions}
\label{supp_sec:discuss}

\subsection{Limitation Discussion}

While $\method$ demonstrates compelling performance in 3DGS personalization conditioned on single-view reference image, several limitations remain. First, the iterative LoRA fine-tuning, though efficient, still requires multiple inference and training rounds, which hinders the application in large-scale batch editing scenarios. 
Besides, faithfully reproducing extremely intricate visual details, including fine-grained patterns and embedded text, remains a challenging aspect for the current LoRA module.
Second, due to the lack of effective automatic 3DGS~\cite{kerbl20233d} insertion method, our method still relies on user-provided bounding boxes in part of scenarios. 
In future work, we plan to develop automatic insertion strategies and enhance scalability by distilling the iterative process into a single forward pass.

\subsection{Potential Societal Impacts}

Our CP-GS framework offers several positive societal implications. By enabling high-quality 3D scene personalization from only a single reference image, it significantly reduces the dependency on extensive image collections or manual 3D modeling, thereby lowering both the cost and workload typically required from human artists. This contributes to a more accessible and efficient content creation process, aligning with the goals of sustainable and green AI.

On the other hand, the ability to easily personalize 3D content may raise concerns regarding potential misuse, such as the generation of inappropriate or harmful scenes involving graphic, violent, or NSFW elements. To mitigate this risk, CP-GS builds upon the diffuser~\cite{von-platen-etal-2022-diffusers} codebase that inherits safety mechanisms—such as NSFW content filters—from the underlying model~\cite{flux2024}. These filters help ensure that the personalized outputs remain within acceptable and responsible usage boundaries.